\documentclass[preprint,12pt,authoryear]{elsarticle}

\usepackage{amssymb}
\usepackage[authoryear]{natbib}
\usepackage{subcaption}

\usepackage{tikz}
\usepackage{float}
\usepackage[edges]{forest}
\usepackage[T1]{fontenc}
\usepackage{lmodern}  
\usepackage[table]{xcolor}
\usepackage{makecell}
\usepackage{multirow}
\usepackage{booktabs}
\usepackage{amsmath}
\usepackage{url}
\usepackage{gensymb}
\usepackage{graphicx}
\usepackage{verbatim}
\usepackage{tabularx} 
\usepackage{algorithm}
\usepackage{algorithmicx}
\usepackage{algpseudocode}
\usepackage{minted}
\usepackage{listings}
\usepackage[utf8]{inputenc}
\RequirePackage{stfloats}
\usepackage{xurl}
\usepackage[colorlinks=true, linkcolor=blue, citecolor=blue, urlcolor=blue, hidelinks]{hyperref} 
\usepackage{adjustbox}

\def\tsc#1{\csdef{#1}{\textsc{\lowercase{#1}}\xspace}}
\tsc{WGM}
\tsc{QE}

\begin{document}

\begin{frontmatter}


\title{An Adaptive Control Architecture for Slope and Terrain Compensation in Autonomous Navigation in Mediterranean Greenhouses}


\author[UAL-Inf]{Fernando Cañadas-Aránega}\corref{corr}\ead{fernando.ca@ual.es}
\cortext[corr]{Corresponding author}

\author[TUM]{Dirk Wollherr}\ead{dw@tum.de}

\author[UAL-Inf]{José L. Guzmán}\ead{mmr411@ual.es}

\author[UAL-Inf]{José C. Moreno}\ead{jcmoreno@ual.es}

\author[UAL-Eng]{José L. Blanco-Claraco}\ead{jlblanco@ual.es}


\affiliation[UAL-Inf]{
    organization={Department of Informatics, CIESOL, ceiA3, Universidad de Almería},
    addressline={Ctra. Sacramento s/n},
    city={Almería},
    postcode={04120},
    country={Spain}
}

\affiliation[UAL-Eng]{
    organization={Department of Engineering, CIESOL, ceiA3, Universidad de Almería},
    addressline={Ctra. Sacramento s/n},
    city={Almería},
    postcode={04120},
    country={Spain}
}

\affiliation[TUM]{
    organization={Chair of Automatic Control Engineering, Technical University of Munich},
    addressline={Arcisstraße 21},
    city={Munich},
    postcode={80333},
    country={Germany}
}


\begin{abstract}
The ability to move stably over terrain with varying slopes and textures is essential for mobile agricultural robots operating in complex and dynamic environments such as greenhouses, where small terrain irregularities can lead to significant navigation errors. This article presents a novel terrain-adaptation strategy based on the carried payload, ensuring accurate and robust trajectory tracking. The proposed approach is based on: (i) the experimental characterization of the most common types of greenhouse soil, concrete, compacted sand, and gravel, and (ii) the direct measurement of terrain slope using the IMU, in order to estimate the force with which this angle affects the motor input. Based on this information, a cascade trajectory-tracking scheme has been designed, consisting of a model-based predictive controller (MPC) in the outer loop and a PI controller in the inner loop. The system incorporates an adaptive feedforward control through gain scheduling approach, capable of adjusting to disturbances caused by variations in slope and terrain type. Simulation results demonstrate that the differential-drive robot achieves a significant improvement both in error indices and in control signal efficiency, highlighting the effectiveness and robustness of the proposed approach.
\end{abstract}



\end{frontmatter}

\section{Introduction}

The rapid growth of the worlds population poses one of the most significant challenges for agriculture: producing more food in less space and sustainably. According to the Food and Agriculture Organisation of the United Nations (FAO), global food demand will increase by 70\% by 2050, driven by both overpopulation and the growing need to feed humans and animals \citep{FAO2024}. This increase is occurring in a context where arable land is becoming increasingly limited, whether due to urbanisation, soil degradation, etc. Against this backdrop, greenhouses are emerging as a strategic and sustainable solution to ensure global food security. These infrastructures enable intensive, controlled food production, optimising the use of available space and reducing dependence on external conditions. Currently, greenhouses cover more than 496,800 hectares globally, of which 42.7\% are located in the Mediterranean basin, a leading region in innovation and agricultural productivity under plastic \citep{zhang2022many}.

However, for greenhouses to consolidate themselves as the true answer to the global food challenge, it is necessary to incorporate advanced technology that maximises their performance and sustainability. Automation, sensorisation, and agricultural robotics are essential for optimising production, reducing labour costs, and improving control over critical variables such as temperature, humidity, and radiation. Nevertheless, the implementation of robots in greenhouse environments remains a complex technical challenge \citep{wang2024path}. The structural conditions of conventional greenhouses are often unfavourable for the operation of mobile robots, especially in complex environments where free space is limited and potential trajectories are obstructed by irregular obstacles. This problem becomes even more pronounced in Mediterranean greenhouses, which are designed to maximise production by minimising uncultivated space, drastically reducing manoeuvring areas and hindering the movement of large robots or those with complex kinematics \citep{canadas2024greenbot}. For this reason, a critical challenge arises when deploying robots in greenhouses, as even small variations in the ground surface can lead to localization losses due to unintended changes in the angular velocity of each wheel. Moreover, given the extreme conditions previously described in a greenhouse, the sensors may not fully compensate for these effects, potentially resulting in navigation failures.

In this context, the different soil types, concrete, compacted sand, or gravel, and the presence of slope variations found in greenhouses can significantly alter the robot’s dynamic states, directly affecting its stability and control capabilities. In \cite{gonzalez2009navigation}, the main characteristics of a greenhouse environment are analysed from the perspective of robotic navigation, highlighting that uneven terrain and difficulties in accurately acquiring sensory data directly affect low-level motor controllers and, consequently, may lead to navigation failure. Several studies, such as \citep{wang2024path}, propose a trajectory-tracking algorithm that emphasises the influence of terrain type and transported payload on navigation performance, demonstrating that ground morphology and payload significantly affect motion control. Likewise, \cite{canadas2024pid} presents methods for estimating and adapting to terrain slope, but points out that their implementation in greenhouses is complex and laborious, especially in environments where the margin for error is very limited. The heterogeneous soil conditions in these environments can prevent the robot from adapting properly, leading to undesirable phenomena such as slippage or unintentional excavation, which represent severe low-level issues that must be resolved to develop reliable high-level algorithms.

This article presents a strategy for adapting to terrain type and slope in order to reduce low-level velocity-tracking errors, thereby decreasing the likelihood of collisions and ensuring stable and robust task execution by reducing the dependence on sensors for navigation corrections. The method is based on the experimental characterization of the most common greenhouse soils and on slope estimation using an inertial measurement unit. With this information, a cascade control architecture is designed for trajectory tracking, incorporating an adaptive FeedForward (FF) controller based on a gain scheduling approach to compensate for disturbances, as well as a Model-based Predictive Controller (MPC) that adjusts the velocity according to terrain conditions, resulting in smaller deviations and thus greater stability. Simulation results show a significant reduction in tracking error and lower control effort, demonstrating the effectiveness of the proposed approach by improving navigation quality in complex environments, especially where the margin for error is smaller.

The structure of the paper is as follows: Section \ref{sec: 2} describes the objective problem in detail. Section \ref{sec: 3} describes the proposed control scheme, as well as the methodology followed to identify the parameters. Section \ref{sec: 4} presents the results and discusses them. Finally, Section \ref{sec: 5} summarises the main contributions.

\section{Experimental Setup} \label{sec: 2}

\subsection{Mediterranean greenhouse}

To perform the simulation results presented in this study, a 3D model of the greenhouse is used, which was developed based on the Agroconnect facilities located in the municipal district of La Cañada de San Urbano (Almería, Spain). A 20 × 20 m model was adapted from the complete greenhouse, consisting of five aisles flanked on both sides by tomato plants, separated by 4 m, forming the robot's navigation corridors. The 3D model accurately replicates a real pear-type tomato crop in a hydroponic system and the geometry of a Mediterranean greenhouse, providing a realistic environment for validating the navigation algorithms.

\begin{figure}[htbp]
\includegraphics[width=0.8\linewidth]{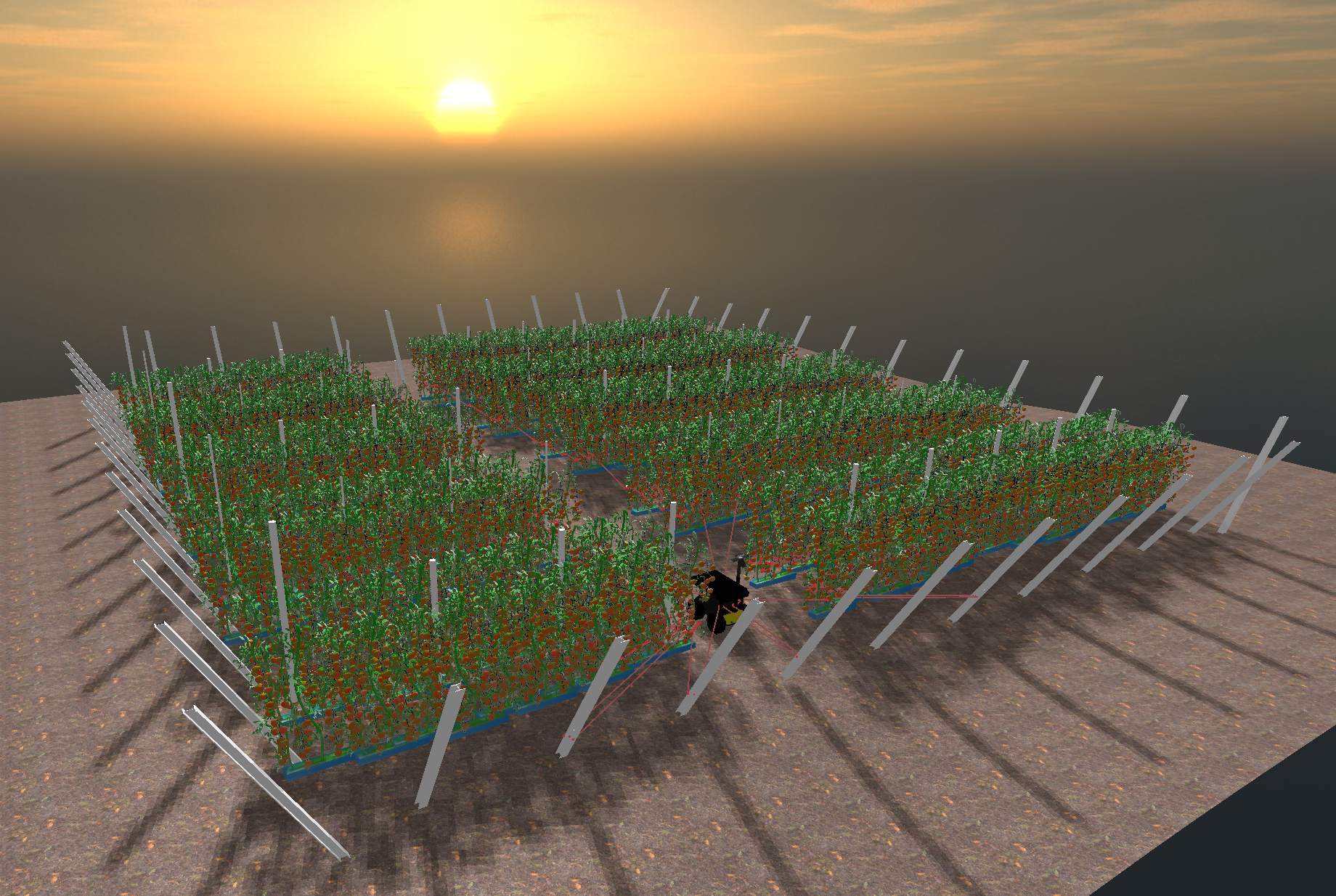} \centering
\caption{3D model of the greenhouse in MVSim} 
\label{fig:Fig2}
\end{figure}

The MultiVehicle Simulator (MVSim) \citep{blanco2023multivehicle} has been used as the simulation environment (see Fig. \ref{fig:Fig2}), as it employs realistic physics-based friction models for tyre-ground interaction, which are ideal for analysing disturbances to develop a control law \citep{canadas2026ros2}. 

\subsection{AgriCobIoT I Robot}

As an autonomous vehicle, the AgriCobIoT I four-wheel mobile robot with skid-steer kinematics is used, based on Clearpath's A200 model, but structurally adapted for performing collaborative agricultural tasks in greenhouses. This system constitutes the leading control platform, whose dynamics are described by the states of a differential robot $\mathbf{x}(t) = [\,x(t)\; y(t)\;\theta(t)\,] \in \mathbb{R}^{3}$, for $t \in \mathbb{R}^{+}$.

For low-level control, the simulator reproduces the dynamics of the engine's viscous friction $b\in \mathbb{R}$ in $\mathrm{N\, m \, rad^{-1} \, s^{-1}}$ (variables taken from the datasheet for the MMP S22-346G-24V GP52-04.3 BR-005 motor), and wheel inertia $J \in \mathbb{R}$ in kg$\,$m$^2$, taking into account the mass of the chassis of the robot $m_{v}\in \mathbb{R}^{+}$ and the wheels $m_w\in \mathbb{R}^{+}$, the sum being the total mass $m_{robot} \in \mathbb{R}^{+}$ in kg. The total $J$ combines the rotational energy of the wheels and vehicle $I_{yy}\in \mathbb{R}^{+}$, together with the effects of dynamic friction and damping $C_D\in \mathbb{R}^{+}$. The resulting dynamics are expressed in equation~(\ref{eq: mvsim22}), where the torque applied by the motors $\tau_{m,i}\in \mathbb{R}$ N$\,$m and their torque constant $k_\tau$ are related to the angular velocity of the drive wheels $\omega_i\in \mathbb{R}$ rad/s, with $i \in \{1, 2\}$ representing each motor.  
\begin{equation}
J\,\dot{\omega}_i(t) + b\,\omega_i(t) = k_\tau\,\tau_{m,i}(t) .
\label{eq: mvsim22}
\end{equation} 
With this equation, the basis for the robot's dynamics model is established, and expanded using a friction model in the following section.

\subsection{Friction simulator model} \label{sec:2.1}

The MVSim friction model interprets the environment as an individual interaction between each wheel of the robot, decomposing the total forces $F_{t,i}(t) \in \mathbb{R}^+$  into longitudinal $F_{log,i}(t) \in \mathbb{R}^+$ and lateral $F_{lat,i}(t) \in \mathbb{R}^+$ components. The simulator evaluates the dynamic friction prior to slippage, establishing a maximum limit per wheel in both directions, $F_{r_{max}} = \mu \, m_{pw}\, g \in \mathbb{R}$, where $\mu \in \mathbb{R}^{+}$ is the coefficient of friction between the wheel and the ground, $m_{pw} \in \mathbb{R}^{+}$ (obtained as $m_{pw}=m_{robot}/nW$, where $nW\in \{1, 2,3,4\}$ is the number of wheels) is the effective mass per wheel in kg, and $g = 9.81\ \text{m/s}^2$ is the gravitational acceleration. The lateral friction is given as $F_{lat,i}(t) = a_{wy,i}(t)\, m_{pw}$, where $a_{wy,i}(t) \in \mathbb{R}$ $\text{m/s}^2$ is the transverse acceleration of the wheel. The longitudinal force, $F_{log,i}(t) = \frac{1}{r} \, (\tau_{m,i} - I_{yy}\, \omega_{{ref,i}} - C_D \, \omega_i)$ (where $r \in \mathbb{R}^{+}$ m is the radius of the wheel), which opposes the direction of travel of the robot, is obtained considering the torque developed by the motor $\tau_{m,i}(t)$, the inertia $I_{yy}$, the reference angular velocity $\omega_{ref,i}(t)$ rad/s and the damping component $C_D$. This force reduces the motor angular velocity $\omega_i(t)$, causing a deceleration proportional to the friction generated.

\subsection{Payload, soil type, and terrain slope}
\label{sec:disturbances}

The MVSim motor dynamics and friction model includes parameters that directly influence the robots behaviour, affecting speed and trajectory control. However, in real conditions, the robot experiences real-time changes while performing its tasks, such as transporting a possible load $m_{pl}\in \mathbb{R}^{+}$, which must be added to $m_{robot}$ and therefore modifies the value of $m_{pw}$, the slope of the terrain $\phi\in \mathbb{R}$, the type of ground on which it performs the tests associated with a change in the values of friction $\mu$, damping coefficient $C_D$ and rolling coefficient $C_{rr} \in \mathbb{R}$  \citep{wong2022theory, fortunato2017dependency}. 

\begin{itemize}
    \item \textit{Load sensitivity}
\end{itemize}

Load sensitivity is associated with an increase in normal force with increasing total system mass, thereby increasing the tyre's contact surface with the ground. This phenomenon reduces the robot's angular velocity by simultaneously affecting the normal force, the moment of inertia, and friction-related forces. In this paper, two conditions are considered: (i) a robot without a payload $m_{pl}=0$ kg, and (ii) a robot with a maximum payload of $m_{pl}=70$ kg (maximum load). From a mathematical perspective, this value varies with the type of transport, so it will be treated as a variable $p \in \{0, 70\}\subset \mathbb{R}$ that changes with the payload, and will be included in the problem formulation.

\begin{itemize}
    \item \textit{Terrain type}
\end{itemize}

\begin{figure}[!ht]
\centering
\includegraphics[width=0.8\linewidth]{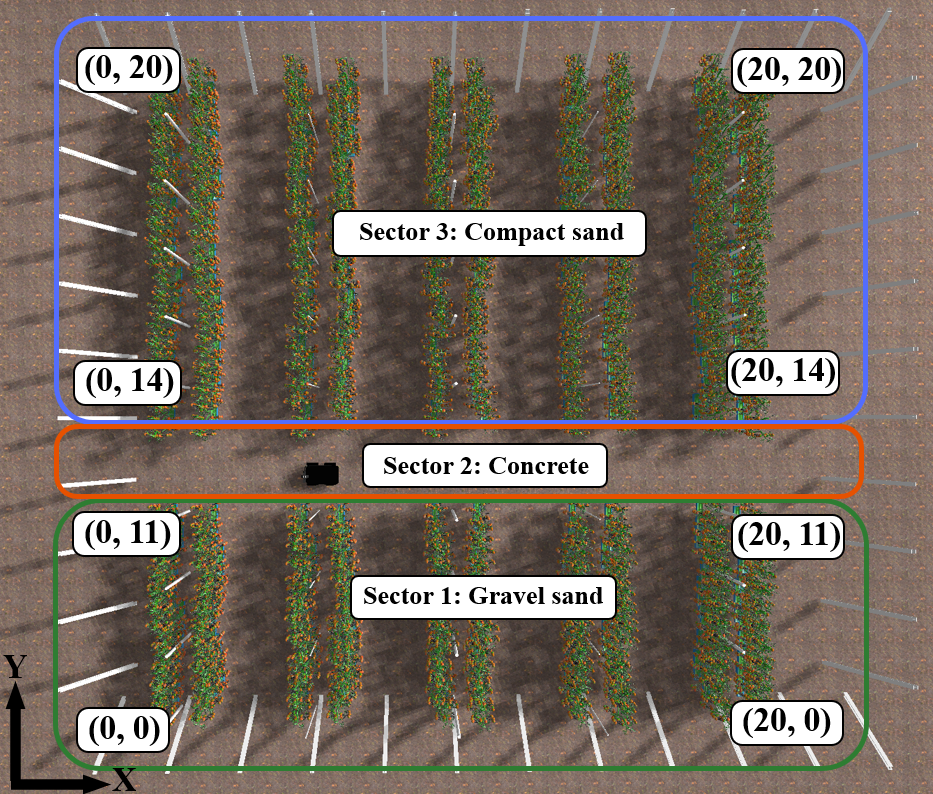}
\caption{Distribution of greenhouse sectors (in metres)}
\label{fig:Fig5}
\end{figure}

Mediterranean greenhouses have different types of soils that often show irregularities and slopes of up to 3-4\,\%, even 5 \%  \citep{canadas2024pid}, affecting speed due to changes in values $\mu$, $C_D$, and $C_{rr}$ in real time. As the values of these parameters change throughout the simulation, another soil-dependent variable $$s \in \{1,2,3\} \subset \mathbb{R}$$ is introduced and updated over time. This value updates the disturbance's influence based on the area where the robot is located. For this reason, in order to replicate the actual conditions of the Agroconnect greenhouse, three types of terrain are defined and distributed by area in the simulation scenario (Fig. \ref{fig:Fig5}). Table~\ref{tab:merged} summarises the $\mu$, $C_D$ and $C_{rr}$ values assigned to each type of terrain, $s$.

The terrain in sector 1, ‘Gravel sand,’ represents a particularly critical environment for autonomous vehicles, as it increases the likelihood of skidding and the safety risk \citep{canadas2024autonomous}. Under such conditions, the EN ISO 3691-4 standard establishes safety requirements for autonomous vehicles when ground conditions are unfavourable (e.g. $\mu < 0.5$), including speed limitation and mandatory braking distance.

\begin{itemize}
    \item \textit{Terrain slope as a disturbance}
\end{itemize}

\begin{figure}[htbp]
\includegraphics[width=0.8\linewidth]{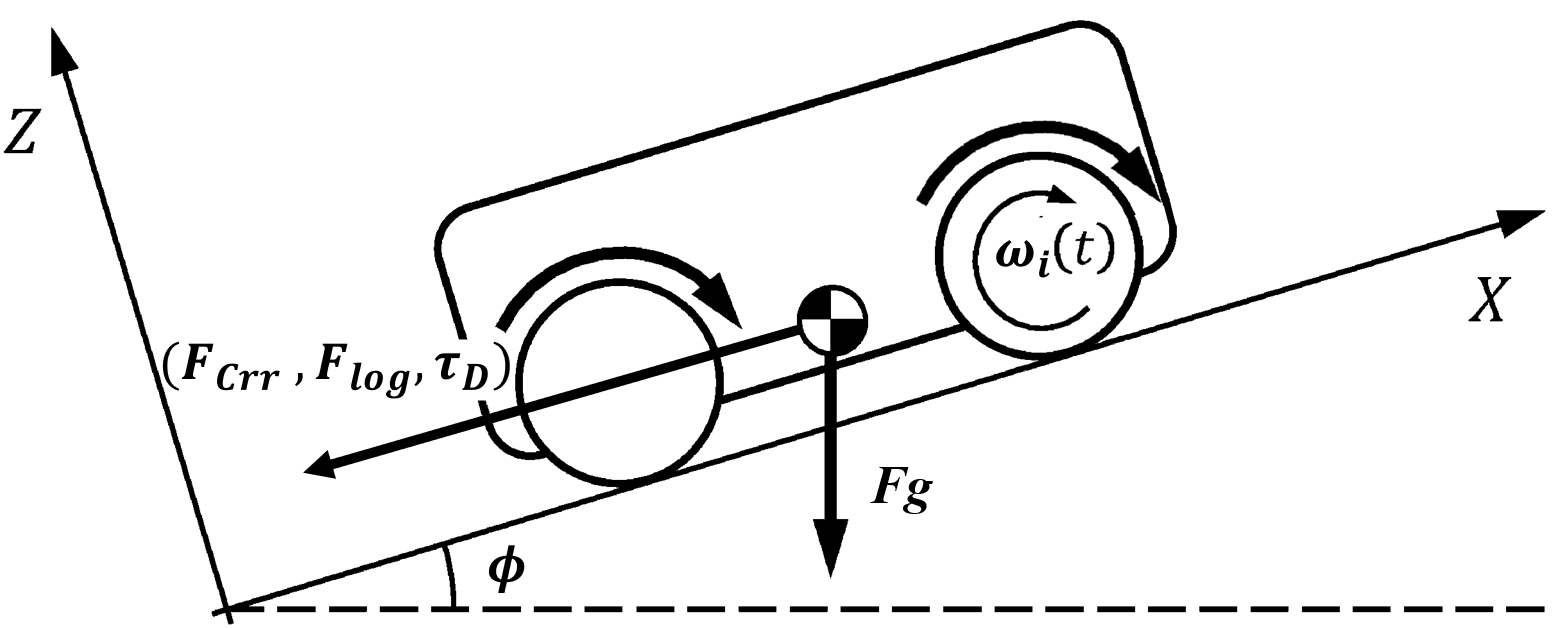} \centering
\caption{Breakdown of forces on sloping terrain} \label{fig:Fig4}
\end{figure}

The presence of slopes, even if they are small, constitutes an external disturbance that directly changes the dynamics of the motor, affecting the angular output speed. On the one hand, gravity attracts the robot's weight in the normal direction of the robot, increasing or decreasing the speed depending on the slope $\phi$. Moreover, other forces (e.g., rolling resistance, gravity force, or torque due to damping) also appear that oppose the direction of the robot and change with $\phi$ \citep{hyon2008compliant} (Fig. \ref{fig:Fig4}). Therefore, a slope introduces a disturbance to the system as an additional resistive torque, whose magnitude depends on the angle of inclination and the robot's total mass. In a completely flat environment, the torque required to maintain a constant angular velocity is limited to overcoming friction and mechanical losses, which are often negligible. However, when a slope is present, the motor experiences a transient or sustained drop in angular velocity due to the additional load.

\section{Proposed adaptive control approach} \label{sec: 3}

\begin{figure*}[htbp]
 \includegraphics[width=\linewidth]{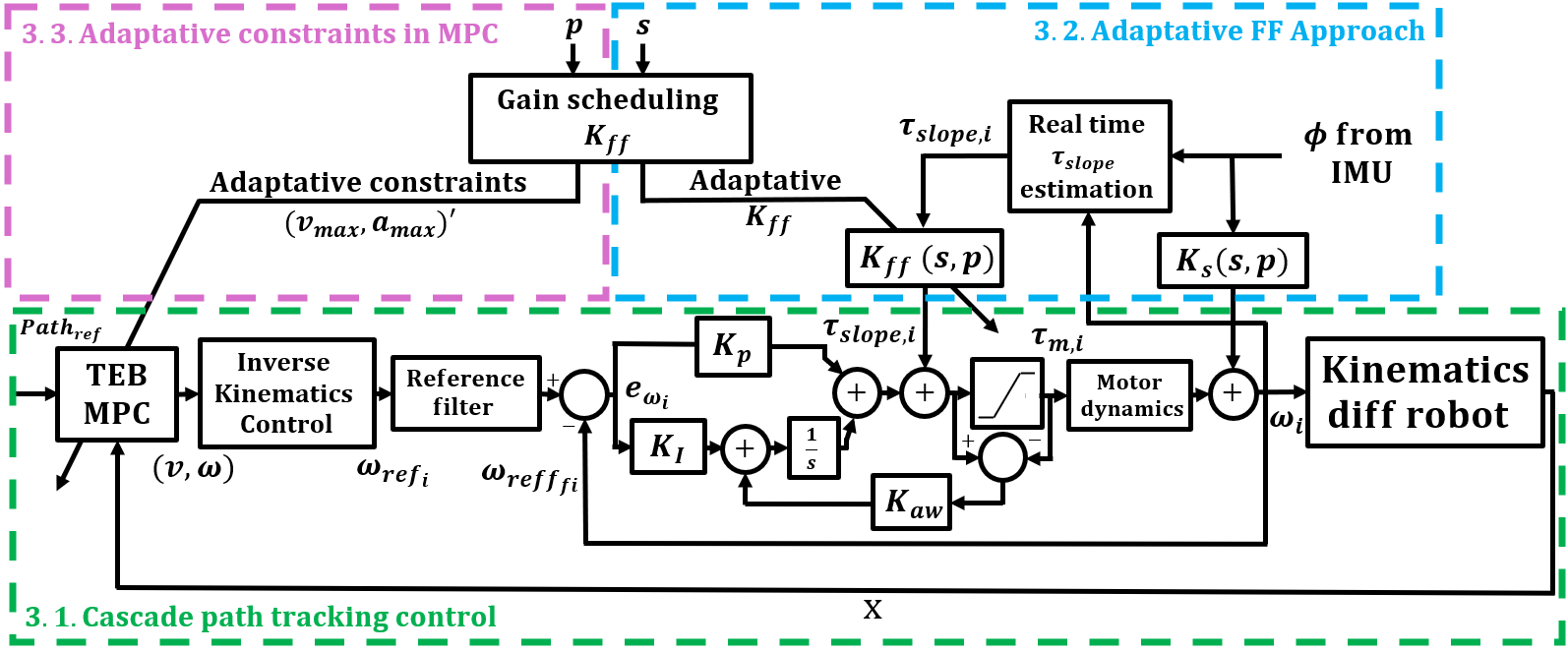} \centering
 \caption{Proposed control architecture} 
 \label{fig:Fig555}
 \end{figure*}

Given the significant influence exerted by payload variation, terrain type, and ground slope on the robot’s dynamics described in the previous section, it becomes necessary to design a control strategy capable of compensating for these disturbances in a robust and real-time manner. This section presents the proposed control architecture, specifically developed to mitigate the combined effects of these external variations on the system. The methodology integrates both the online estimation of environment-dependent parameters and an adaptive FF controller to compensate for their effects. The proposed control architecture is summarized in Fig. \ref{fig:Fig555}. The control scheme is based on a cascade approach, with an inner loop dedicated to motor speed control using PID controllers and an outer loop centered on MPC, designed to address trajectory tracking and compute the optimal motor speed references. The control scheme is completed with the main contribution of this work, which is the estimation of the force acting on the wheels due to terrain slope, together with a gain-scheduling algorithm to adapt a FF compensator gain according to the disturbances.

\subsection{Cascade path tracking control}
This section summarizes the design of the cascade control loops, where the values for the different parameters are summarized in Table \ref{tab:2}. For the design of the internal loop, data from AgriCobIoT I (see Section 2.2) is obtained to calculate the robot's motor dynamics. The parameters $b$ and $k_\tau$  are used in (\ref{eq: mvsim22}) to obtain the corresponding first-order model with a static gain  $K_m \in \mathbb{R}$ and a time constant $\tau \in \mathbb{R}$. Using this linear model, PI controllers with parameters $K_P$ and $K_I$ were designed for the speed control, with a closed-loop specification given by two poles at $-0.6$ and $-31.25\,\mathrm{s^{-1}}$ ~\citep{ast+haggISA2006}. The PID control loops were extended with a reference filter with time constant, $\tau_f$, to reduce overshoot in the closed-loop response, and a back-calculation control scheme  with a tracking constant $K_{aw} \in \mathbb{R}$ to deal with the the torque saturation limits  $[-25,25]\,\mathrm{N\,m}$. For the external loop control, a MPC based on the Timed Elastic Band (TEB) is used, including as constraints the maximum robot longitudinal velocity, $v_{\max}$, the maximum acceleration, $a_{\max}$, a minimum distance to obstacles, $d_{\text{obst}}$, and using the const function weighting matrices $Q$ and $R$ (see \citep{rosmann2015timed} for more details).

\begin{table}[h]
\centering
\caption{Cascade path tracking control params}
\begin{tabular}{l c l c}
\hline
\textbf{Params} & \textbf{Value used} & \textbf{Params} & \textbf{Value used} \\
\hline
$J$ & 2.22 kg$\, $ m$^{2}$ & $K_P$ & 70 N$\, $m$\, $s\\
$b$ & 0.75 N$\, $m$\, $s$\, $rad$^{-1}$ & $K_I$ & 40 N$\, $m \\
$K_m$ & 1.33 N$\, $m$\, $s$\, $rad$^{-1}$ & $K_{aw}$ & $K_p/K_I$ s$^{-1}$ \\
$\tau$ & 2.97 s & $\tau_f$ &  $K_p/K_I$ s\\ \hline
$v_{\max}$ & 0.75 m/s & $d_{\text{obst}}$ & 0.5 m\\
$a_{\max}$ & 2 rad/s$^{2}$ & $Q$, $R$ & $50I$, diag(0.5,\,1.0)\\
\hline
\end{tabular}
\label{tab:2}
\end{table}



\subsection{Adaptative FF Approach}
Once the cascade control is designed, this section introduces an online estimate of the force acting on the wheels due to terrain slope, soil type, and robot payload, which is used to design a FF compensator whose objective is to counteract, in real time, the effects of terrain variations on the motor dynamics and, consequently, on the controlled angular velocity of the robot.

The following forces are taken into account to estimate this value: dynamical friction $F_{r_{max}}(s,p)= \mu(s) \, m_{pw} (p) \, g$, rolling resistance $F_{C_{rr}}(s,p,\phi)= C_{rr}(s) \, m_{pw} (p) \, g \, \cos{(\phi)}\in \mathbb{R}$, gravitational force $F_g(p,\phi)=m_{pw} (p) \, g \, \sin{(\phi)}\in \mathbb{R}$ and damping torque $\tau_{D}(s,\omega_i)= C_D(s) \, \omega_i \in \mathbb{R}$ (shown in the Fig. \ref{fig:Fig4}), where $s$ represents the soil type, and $p$ the robot payload, previously described.  Table~\ref{tab:merged} summarises the different values of $m_{pl}$, $\mu$, $C_D$ and $C_{rr}$ based on soil type and payload. The terrain slope angle $\phi(t)$ is measured in real time from the tilt value provided by the MVSim simulator using the Mobile Robot Programming Toolkit (MRPT) library. This tool uses the \texttt{getPose3D} message, which calculates the orientation from the robot's odometry and simulated information, emulating a real IMU with magnetometers.

The result of the slope torque estimation $\tau_{slope}(s,p,\phi,\omega_i) \in \mathbb{R}$ affecting the motor input is represented in equation (\ref{eq:torque_slope}), evaluating the longitudinal force $F_{slope}(s,p,\phi)\in \mathbb{R}$ (\ref{eq:F_rmax}).
\begin{align}
F_{slope}(s,p,\phi) =
\begin{aligned}[t]
\operatorname{max}\{ -F_{r_{\max}}(s,p),\,
&\operatorname{min}\{ F_g(p,\phi)+ \\ + F_{Crr}(s,p,\phi), 
& F_{r_{\max}}(s,p)\}\} ,
\end{aligned}
\label{eq:F_rmax} \\
\tau_{slope}(s,p,\phi,\omega_i) = F_{slope}(s,p,\phi)\, r - \tau_{D}(s,\omega_i).
\label{eq:torque_slope}
\end{align} 
Once a real-time estimate of the force acting on the robot’s wheels is available, several simulations were carried out with the aim of quantifying the impact of the slope-induced disturbance on motor velocity. To this end, a series of tests was conducted in which the system was driven to a steady-state linear velocity, allowing the static relationship between slope, the torque variation caused by the slope, and the resulting change in motor output to be observed. This procedure enables the identification of the static gain, denoted as $K_{s}(s,p) \in \mathbb{R}$ and expressed in $\mathrm{rad\, s^{-1} \, N^{-1}\, m^{-1}}$, which is defined as the variation in the process output, $\Delta \omega_i$, with respect to the variation of the disturbance influence on the input torque, $\Delta \tau_{slope}$, associated with the angle $\phi$. The decision to relate $\tau_{slope}$ to $\omega_i$ for computing $K_s(s,p)$ allows the direct analysis of the slope’s effect on the system input, resulting in more accurate values. These simulations were repeated for different payload conditions, $p$, and for the various terrain types, $s$, considered in this study, as summarized below:

\begin{itemize} 
    \item \textit{Test 1 -- Terrain slope}: Keeping the soil type constant, the robot moves on slopes between (-5,5)\%, obtaining the corresponding  $K_{s}$ gains for each case. This process is repeated for a maximum payload $m_{pl}=70$ Kg.
    
    \item \textit{Test 2 -- Terrain slope and terrain change}: in the same way, a second test is carried out where a variable slope is combined with a change terrain type (different sectors in Figure \ref{fig:Fig5} and, therefore, changing $s$) and maximum payload, repeating the test for $m_{pl}=70$ kg.
\end{itemize}

Based on this information, a FF compensator for each value of $K_s(s,p)$ is calculated \citep{guzman2024feedforward}, with a non-dimensional static FF gain of $K_{ff}(s,p)=K_s(s,p)/K_m\in \mathbb{R}$, with $K_m = 1.33\,\mathrm{N\, m \, rad^{-1}\, s^{-1}}$ such as defined in Table \ref{tab:2}. This value weights the estimated torque slope $\tau_{slope}(s,p,\phi,\omega_i)$ to act on the system input. As a result, a gain-scheduling approach to update the FF gains based on the different values of payload $p$, and the terrain type $s$ is designed and summarized in Table~\ref{tab:merged}.


\begin{table}[ht]
\centering
\caption{Combined gain-scheduling table for $K_{ff}$ with friction parameters \citep{wong2022theory}}
\resizebox{\columnwidth}{!}{%
\begin{tabular}{lccccccc}
\hline
\textbf{Terrain - $(s)$} & \textbf{Sector} & \textbf{$m_{pl}$ - $(p)$ [kg]} &
$\boldsymbol{\mu}$ & $\boldsymbol{C_\mathrm{rr}}$ & $\boldsymbol{C_D}$ & \textbf{$K_{ff}$} \\
\hline
\multirow{2}{*}{Gravel sand} 
 & \multirow{2}{*}{1} & 0   & \multirow{2}{*}{0.2} & \multirow{2}{*}{0.10} & \multirow{2}{*}{1.25} & 0.031 \\
 &  & 70  &  &  &  & 0.025 \\ 
\hline
\multirow{2}{*}{Concrete} 
 & \multirow{2}{*}{2} & 0   & \multirow{2}{*}{0.8} & \multirow{2}{*}{0.01} & \multirow{2}{*}{0.75} & 0.021 \\
 &  & 70  &  &  &  & 0.051 \\ 
\hline
\multirow{2}{*}{Compact sand} 
 & \multirow{2}{*}{3} & 0   & \multirow{2}{*}{0.5} & \multirow{2}{*}{0.05} & \multirow{2}{*}{1.00} & 0.021 \\
 &  & 70  &  &  &  & 0.015 \\ 
\hline
\end{tabular}}

\label{tab:merged}
\end{table}

 
\subsection{Adaptative constraints in Model Predictive Control}
\label{sec:MPC}
 Due to friction, there are some occasions where the longitudinal force exerted $F_{{long}}$ exceeds the maximum limit $F_{r_{max}}$ depending on the sector, causing the robot to swerve and potentially threatening the environment and farmers. Therefore, the maximum value of the constraints $v_{max}$ and $a_{max}$ in the MPC algorithm is modified in real time by a value $\beta_{mpc}=\frac{F_{r_{max}}} {F_{{long_{max}}}}\, 0.9\in \mathbb{R}$, where $0.9$ corresponds to 10\% as the safety coefficient of the ISO~3691-4 standard, adapting the maximum value of velocity $v_{max}'=v_{max}\, \beta_{mpc}$ and acceleration $a_{max}'=a_{max}\, \beta_{mpc}$.

\section{Results} \label{sec: 4}

\begin{figure*}[htbp]
\includegraphics[width=1\linewidth]{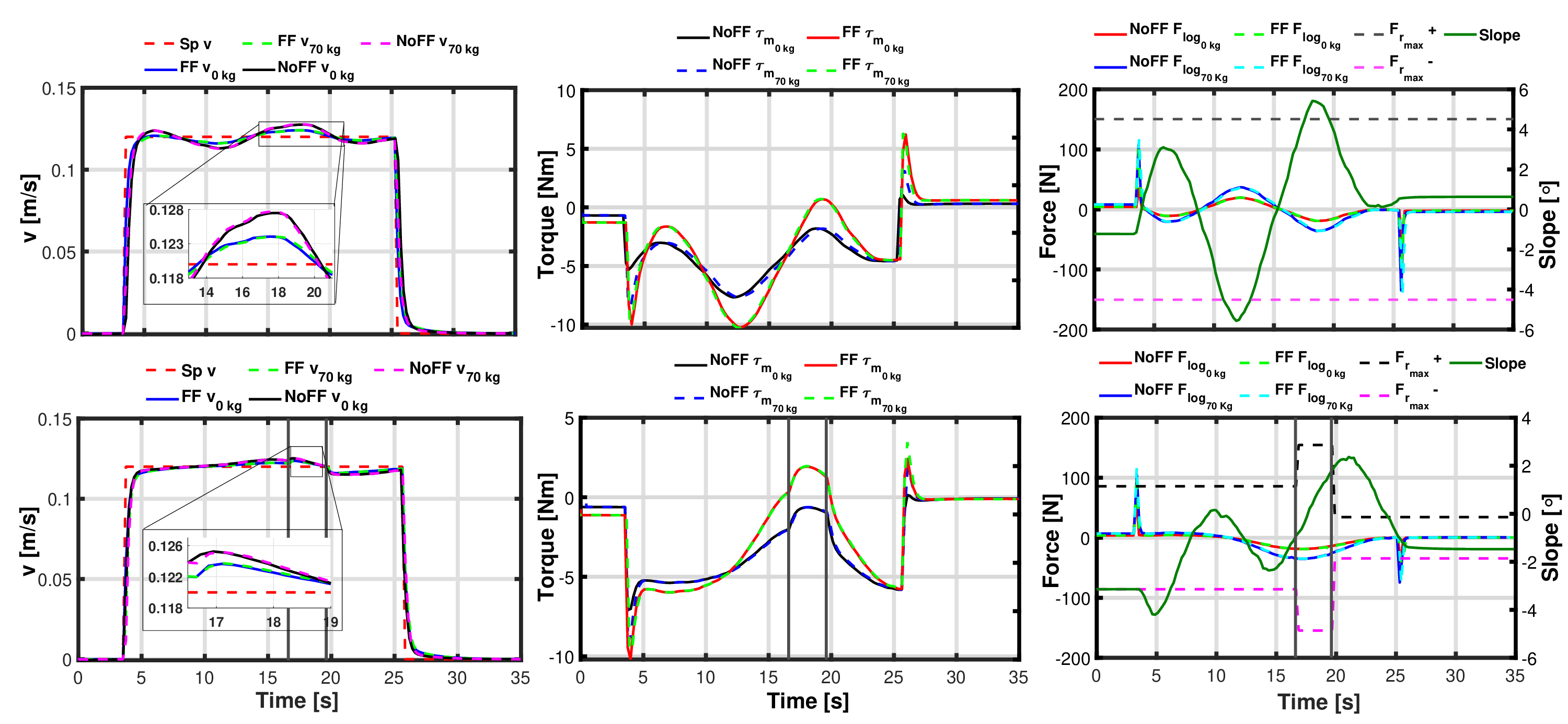} \centering
\caption{First row for test 1 (same terrain) and second row for test 2 (different terrain types). The black vertical lines indicate a change from sector 3 to sector 2 at time 16.50 s and a change from sector 2 to sector 1 at time 19.75 s.} 
\label{fig:Fig55}
\end{figure*}

This section presents the simulation results obtained for the cascade control architecture described in the previous sections, considering both cases, with and without the adaptive FF compensator. The simulation was run on a personal computer equipped with an Intel Core i7-13400K processor, 32 GB of RAM and an NVIDIA GTX 4060 graphics card. A sampling interval of 0.01 s was used for a forecast horizon of 5 time steps. The graphical results are shown in Figure~\ref{fig:Fig55}. Two simulations were performed by driving the robot through the sectors defined in Figure~\ref{fig:Fig5}: (i) traversing only Sector~2, with a step change in the velocity reference to $v = 0.12\,\text{m/s}$, in order to analyze the robot's response under different slope conditions; and (ii) crossing all three sectors using the same velocity profile to assess the combined influence of slope variations and soil-type transitions. In addition, an analysis of the longitudinal forces and the robot’s maximum achievable friction is included.
Table~\ref{tab:IAE} provides a quantitative comparison between the controllers with and without feedforward compensation, reporting the Sum of Absolute Error $(SAE = \sum_{0}^{T} |e_{\omega_i}(k)|)$ and the Sum of Control Increment $(SCI = \sum_{0}^{T} |\tau_{m,i}(k)|)$. The last two columns summarize the corresponding improvement indices achieved by the proposed approach including adaptive feedforward control.

\begin{table}[!h]
\centering
\caption{Quantitative results from SAE - SCI}
\resizebox{\columnwidth}{!}{%
\begin{tabular}{cccccccc}
\hline
\textbf{Test} & \textbf{$m_{pl}$ - $(p)$ [kg]} & \textbf{$SAE_{NoFF}$} & \textbf{$SCI_{NoFF}$} & \textbf{$SAE_{FF}$} & \textbf{$SCI_{FF}$} & \textbf{$Imp.$ $SAE$} & \textbf{$Imp.$ $SCI$}\\ \hline
\multirow{2}{*}{1} & 0  & 0.53 & 58.30 & 0.41 & 59.48 & 22.64 \% & -2.02 \%\\
 & 70 & 0.88 & 75.46 & 0.76 & 75.85 &  13.64 \% & -0.51 \%\\ \hline
\multirow{2}{*}{2} & 0  & 0.53 & 53.30 & 0.44 & 53.95  & 16.98 \% & -1.22 \%\\
 & 70 & 0.63 & 62.13 & 0.57 & 62.95 & 9.52 \% & -1.32 \% \\ \hline
\end{tabular}%
}
\label{tab:IAE}
\end{table}

\begin{figure}[htbp]
\includegraphics[width=\linewidth]{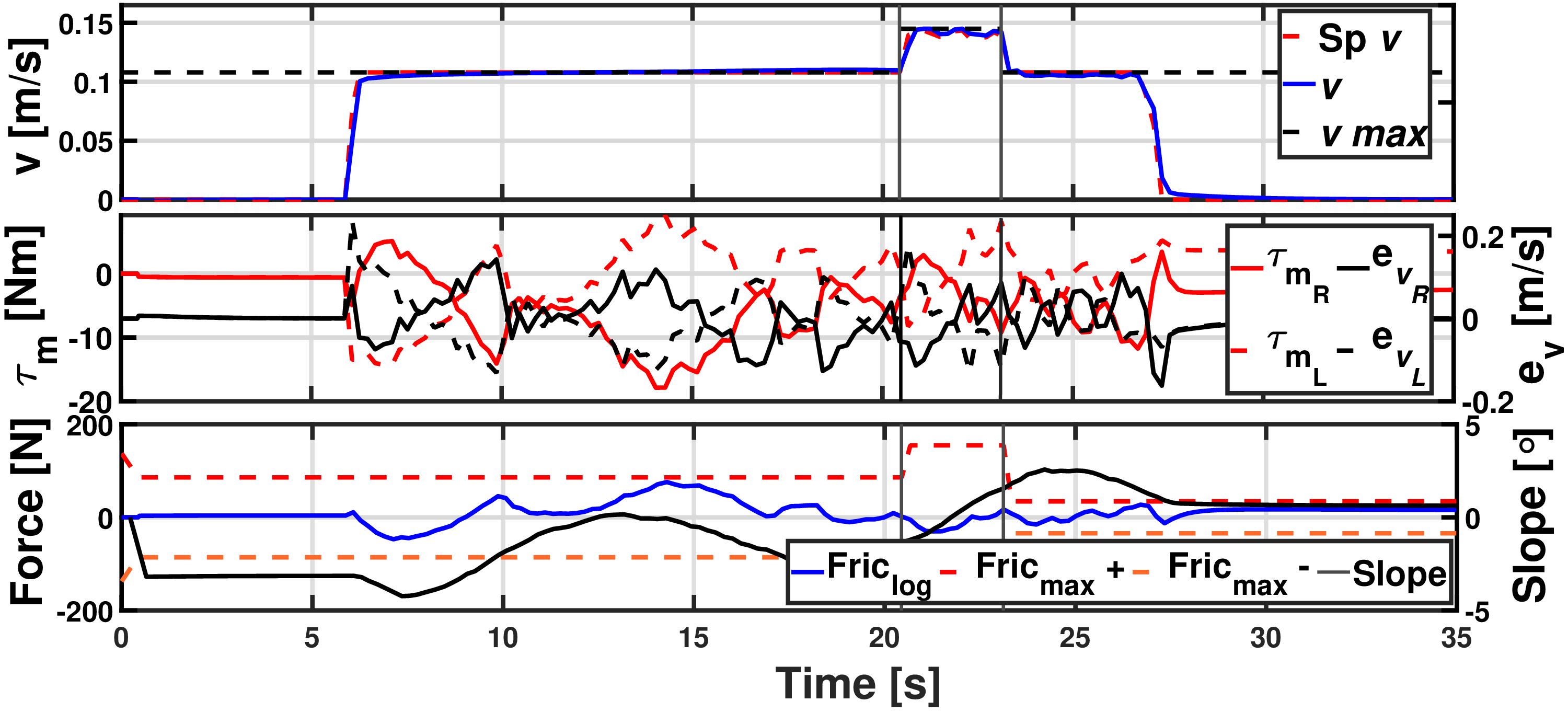} \centering
\caption{Results with constrains including the safety margin} 
\label{fig:Finalmpc}
\end{figure}

As observed in the output response, the proposed approach introduces noticeable performance improvements. Although the enhancement in sector transitions is modest, the slope compensation exhibits a clearer improvement, particularly in the simulations where both perturbations are present. Consequently, the control effort increases slightly due to the action of the FF, in exchange for a considerable reduction in error and, therefore, an improvement in navigation. Finally, the friction analysis reveals that when the robot begins to move, the longitudinal force $F_{log}$ exceeds the admissible limit $F_{r_{\max}}$. In a real system, this would correspond to wheel slip during acceleration or braking. To solve this problem, a trajectory tracking test is analyzed using the constraint adaptation approach proposed for the MPC algorithm. In this case, to determine how much speed must be reduced to comply with regulations in industrial environments, the maximum value of $F_{log}$ is obtained in hazardous areas, adding a 10\% \ safety margin as indicated by the regulations. Figure \ref{fig:Finalmpc} shows the new results with a 17.4\% reduction in the friction and satisfying the constraints. This adjustment prevents any violation of the friction threshold, thereby ensuring that the robot operates in compliance with the specified safety  requirements.

\section{Conclusion} \label{sec: 5}

This work presents an adaptive control strategy based on gain scheduling that significantly improves the trajectory tracking of mobile robots under different payloads, terrain types, and slopes. The combination of an adaptive feedforward compensator and an MPC allows real-time adjustment of the robot’s control dynamics, preventing loss of localisation and navigation failures on heterogeneous soils, thereby reducing the risk of unsuccessful navigation.

The results show that the differential-drive robot achieves significant reductions in tracking error within the confined greenhouse test environment, demonstrating the effectiveness and robustness of the proposed approach. When extrapolated to typical Mediterranean greenhouses, whose operational areas may span several hectares, the improvement becomes even more relevant, as this method makes the robot more accurate and therefore safer. A significant enhancement is also observed at the low-level control stage, reducing the SAE during trajectory tracking. Although feedforward control may slightly increase the SCI, this increase is minimal compared to the substantial gains in precision and stability. 

In practice, the robot can navigate more reliably in narrow spaces and under adverse conditions typical of Mediterranean greenhouses, something that was previously not possible.

\section*{Acknowledgments} 
This work has been carried out within the framework of the LIFE-ACCLI\\AMTE project (LIFE23-CCAES-LIFE-ACCLIMATE/101157315), and a Er-\\asmus+ Mobility grants from CeiA3. The first author, Fernando Cañadas-Aránega, holds an FPI grant (PRE2022-102415) from the Spanish Ministry of Science, Innovation, and Universities..
\bibliographystyle{cas-model2-names}

\bibliography{cas-refs}




\end{document}